\documentclass[acmlarge]{acmart}
\usepackage{color,array}

\usepackage{graphicx}

\usepackage{mathrsfs}
\usepackage{bookmark}
\usepackage{amsmath,bm}

\usepackage{tabularx}
\usepackage{lineno}
\usepackage{epsfig}
\usepackage{chngpage}
\usepackage{float}
\usepackage{graphicx}
\usepackage{array}
\usepackage{diagbox}
\usepackage{epstopdf}
\usepackage{algorithm}
\usepackage{algorithmicx}
\usepackage{algpseudocode}
\usepackage{threeparttable}
\usepackage{dsfont}
\usepackage{subfigure}
\usepackage{multirow}
\usepackage{longtable}
\usepackage{booktabs}
\usepackage{multirow}
\usepackage{amssymb}
\usepackage{booktabs}
\usepackage{amsthm}
\usepackage{url}
\usepackage{footnote}
\makesavenoteenv{table}

\AtBeginDocument{%
	\providecommand\BibTeX{{%
			\normalfont B\kern-0.5em{\scshape i\kern-0.25em b}\kern-0.8em\TeX}}}

\setcopyright{none}
\renewcommand\footnotetextcopyrightpermission[1]{}
\setcopyright{none}
\makeatletter
\renewcommand\@formatdoi[1]{\ignorespaces}
\makeatother

\begin{document}
	
	\title{Brain-inspired Artificial Intelligence: A Comprehensive Review}
	
	\author{Jing Ren}
  \email{ch.yum@outlook.com}	
  \author{Feng Xia}
  \authornote{Corresponding author}
  \email{f.xia@ieee.org}  
	\affiliation{%
		\department{School of Computing Technologies}
		\institution{RMIT University}
		\city{Melbourne}
		\state{VIC}
		\postcode{3000}
		\country{Australia}
	}
	
	\renewcommand{\shortauthors}{Ren and Xia}
	
\begin{abstract}
	

Current artificial intelligence (AI) models often focus on enhancing performance through meticulous parameter tuning and optimization techniques. However, the fundamental design principles behind these models receive comparatively less attention, which can limit our understanding of their potential and constraints. This comprehensive review explores the diverse design inspirations that have shaped modern AI models, i.e., brain-inspired artificial intelligence (BIAI). We present a classification framework that categorizes BIAI approaches into physical structure-inspired and human behavior-inspired models. We also examine the real-world applications where different BIAI models excel, highlighting their practical benefits and deployment challenges. By delving into these areas, we provide new insights and propose future research directions to drive innovation and address current gaps in the field. This review offers researchers and practitioners a comprehensive overview of the BIAI landscape, helping them harness its potential and expedite advancements in AI development.
 
\end{abstract}



	\maketitle
	
\section{Introduction}
A fundamental goal of artificial intelligence (AI) is to create machines that can learn and think like humans. In pursuit of this goal, artificial learners have achieved remarkable milestones across various domains, including object and speech recognition~\cite{prabhavalkar2023end,serban2020adversarial}, image processing~\cite{minaee2021image}, robotics~\cite{garaffa2021reinforcement}, medical data analysis~\cite{suganyadevi2022review}, natural language processing (NLP)~\cite{min2023recent}, and more. These successes have accelerated the progress of AI to the point where it can rival and even surpass humans in certain areas. For instance, AI models now outperform humans in specific tasks such as language translation~\cite{ranathunga2023neural}, image recognition~\cite{he2016deep}, and even strategic games like chess and Go~\cite{silver2016mastering}. More recently, a new family of multimodal models, capable of image, audio, video, and text understanding similar to those of humans have been proposed by many companies~\cite{team2023gemini,achiam2023gpt,anthropic2024claude}. This rapid progress underscores AI's transformative potential across diverse fields, pushing the boundaries of what technology can achieve. However, general AI approaches, which aim to create machines capable of human-like thought and reasoning, still have limitations in terms of scalability, robustness, energy efficiency, interpretability, learning efficiency, and adaptability~\cite{liu2018artificial}.

Human brain, which is acknowledged as the most sophisticated information processing system, is capable of solving complex tasks such as learning, reasoning, and perception. Based on recent advancements in the study of the human brain, researchers are integrating neuroscience insights into AI systems. They aim to develop Brain-Inspired Artificial Intelligence (BIAI) systems that can perceive, reason, and act in ways more akin to human behavior~\cite{peng2024adaptive,sun2024eeg}. This endeavor is rooted in the desire to understand the underlying principles of biological intelligence and to harness them for building more intelligent, adaptive, and robust AI systems. 

\textit{What is BIAI?}
BIAI refers to AI systems and algorithms that take inspiration from the biological structure, function, and principles of the human brain and neural system. It focuses on replicating or imitating the complex processes and functionalities observed in biologies to achieve more human-like or brain-like behavior in artificial systems~\cite{yang2018survey}. Compared with general AI algorithms, BIAI typically concentrates on specific aspects of human behavior, such as learning from experience, adapting to new environments, and paying attention to important information. 

In this comprehensive review, the literature of BIAI is roughly divided into physical structure (PS)-inspired and human behavior (HB)-inspired models. PS-inspired models refer to models imitating the structure of biological neurons, synapses, and neural circuits to perform tasks such as learning, reasoning, and decision-making. Representative models include Multi-layer Perceptron (MLP), Artificial Neural Networks (ANNs), and more recently, Spiking Neural Networks (SNNs). HB-inspired models are defined as models that replicate the biological mechanisms and processes observed in human behaviors. These models aim to capture the dynamics of biological systems while providing insights into how human perceive, learn, adapt, and interact with the environment. Attention mechanism, transfer learning, and reinforcement learning are common deep learning methods inspired by human behaviors.

\textit{Differences between BIAI and general AI} are manifested in different approaches and goals within the field of AI~\cite{chen2022far,kasabov2019time}. Specifically, general AI is not necessarily inspired by the specific workings of the human brain but aims to achieve or even surpass human-level intelligence in a broader sense. On the contrary, the aim of designing BIAI system is to replicate or mimic the biological mechanisms and processes underlying human cognition. These systems generally excel in tasks such as image recognition and robotic control, but they may not necessarily possess the full spectrum of human intelligence. A more comprehensive comparision between BIAI and traditional AI is shown in Table~\ref{difference}.

\textit{Why is BIAI significant?}
BIAI holds significant importance mainly from two perspectives. On the one hand, BIAI has the potential to outperform traditional AI approaches in many aspects, including adaptability, generalization, and interpretability. On the other hand, BIAI models aim to mimic the brain's structure and function, thereby increasing their biological plausibility. This alignment with biological principles not only deepens our scientific understanding of intelligence but also fosters new opportunities for collaboration between neuroscience and AI research. In essence, by drawing inspiration from the human brain—the most advanced information processing system—researchers are setting the stage for the development of intelligent systems that could potentially match or even exceed human-level capabilities~\cite{lu2018brain,parhi2020brain,fan2020brain}.
\begin{table}[htbp]
	\caption{\label{difference}Differences between brain-inspired AI and traditional AI}
	\begin{tabular}{|p{3cm}<{\centering}|p{6cm}<{\centering}|p{6cm}<{\centering}|}
		\toprule
		\textbf{Aspect} & \textbf{Brain-inspired AI}  & \textbf{Traditional AI}\\
		\midrule
		Learning Approach & Mimics human brain learning (e.g., neural networks)& Rule-based, predefined algorithms  \\	\hline
		Adaptability & High, capable of learning and adapting from experience&	Low, limited to programmed instructions\\	\hline
		Efficiency & Optimized for energy-efficient computations&	Often requires significant computational power\\	\hline
		Flexibility & Capable of handling unstructured data and environments&	Struggles with unstructured data and environments\\	\hline
		Intelligence & Emulates cognitive processes and perception&	Follows logical reasoning and statistical methods\\	\hline
		Neuroscience Integration&Directly inspired by brain functions and structures&	Based on mathematical models and data analysis \\	\hline
		Autonomy & High, capable of autonomous decision making&	Limited, requires human intervention for complex decisions\\	\hline
		Learning Mechanism & Continuous learning, self-improving&	Static learning, needs retraining for updates\\	\hline
		Example Applications & Advanced robotics, autonomous systems, cognitive computing&	Traditional data analysis, rule-based systems\\	\hline
		Development Focus & Understanding and replicating brain-like behavior&	Enhancing algorithmic efficiency and accuracy \\	

		\bottomrule
	\end{tabular}
\end{table}

\subsection{Motivation}
The human brain is the pinnacle of biological complexity. It not only regulates all bodily functions and processes but also enables advanced cognitive abilities such as thinking, memory, and emotion~\cite{bassett2011understanding}. Integrating neuroscience with AI system can help solve many pressing issues and certain bottlenecks in various real-world applications~\cite{zhang2020system}. On the one hand, the human brain is remarkably efficient in processing vast amounts of information while consuming relatively low amounts of energy. Mimicking its architecture and processes can lead to AI systems that are similarly efficient and elegant in their operations. For example, traditional robots cannot acquire timely environmental knowledge in complex environments, which limits its capacity of making accurate and fast decisions. Additionally, issues such as low learning efficiency, poor generalization, difficulty in developing goal-oriented strategies, and slow adaptation to dynamic environments still persist in this area. Integrating BIAI into robotic systems can dramatically improve robotic motion and manipulation capabilities~\cite{qiao2021survey}. Moreover, BIAI can also be applied into solving many other real-world problems, such as medical diagnostics, self-driving cars, chatbots and virtual assistant, cyber threat detection, tutoring systems, supply chain optimization, content creation, and personalized recommendations. These applications highlight the broad impact and relevance of BIAI in different aspects.

On the other hand, understanding the brain's mechanisms not only provides insights into how intelligence emerges but also offers clues for solving complex problems in AI. By studying biological neural networks, researchers can develop algorithms and architectures that better capture the intricacies of cognition and perception. For instance, neural networks, one of the foundational and fundamental models in AI, draw inspiration from the brain's structure and computational processes. As a cornerstone of modern AI, neural networks power advancements across healthcare, finance, transportation, and entertainment. Their capacity for learning from data and uncovering valuable insights makes them essential for tackling complex challenges and driving AI innovation. Additionally, the human brain is remarkably robust and adaptable, capable of learning from experience, handling noisy and uncertain data, and generalizing knowledge to new situations~\cite{dosenbach2007distinct}. By mimicking the brain's resilience and adaptability, BIAI seeks to create more robust and versatile AI systems. This approach also prioritizes ethical AI development by emphasizing transparency, interpretability, and accountability. Modeling intelligence on biological systems can foster the creation of AI that is trustworthy and aligned with human values.

While BIAI holds tremendous promise for advancing AI and robotics~\cite{liu2024multiple}, it also presents several challenges and limitations. The human brain is an extraordinarily complex organ with billions of neurons and trillions of synapses, organized into intricate networks that govern cognition, perception, and behavior. Replicating this level of complexity in ANNs poses significant computational and engineering challenges~\cite{strukov2019building}. Due to such complexity of human brain, our understanding of the brain remains incomplete in spite of decades of research. Many aspects of brain function, such as learning, memory, and consciousness, remain poorly understood~\cite{seth2022theories}. This lack of understanding complicates efforts to translate insights from neuroscience into practical algorithms and architectures for BIAI. In addition, the complexity and opacity of BIAI models hinder our ability to understand their decision-making processes. This notable lack of interpretability and transparency raises significant concerns about accountability, bias, and trustworthiness in AI systems~\cite{kaur2022trustworthy,li2023trustworthy}, especially in safety-critical applications like healthcare and autonomous vehicles.

These shortcomings motivate us to conduct a comprehensive study on BIAI. In the literature, several survey papers have investigated the algorithms of BIAI in different application scenarios and from different perspectives. However, most of these studies focus on one specific aspect, such as algorithms, application scenarios, or cost functions, being lack of a comprehensive review of detailed introduction and discussion about current BIAI research progress. In this review paper, we categorize and review current BIAI research based on the inspiration source of algorithms and learning mechanism. For each BIAI algorithm, after introducing its features and suitable applications, we discuss its advantages and disadvantages. Then, we discuss the open problems of current BIAI models and list several future directions. We hope this comprehensive review can provide useful insights to researchers in relevant fields.

\subsection{Related Surveys and Novelty}
Previous works cover similar topics within the scope of brain-inspired/brain-like learning or computing~\cite{schmidgall2024brain,qiao2021survey,jiao2022new,hassabis2017neuroscience}, but none focused on the specific knowledge that neuroscience brought to AI models and introduced BIAI systems comprehensively and in detail. In~\cite{qiao2021survey}, the authors seek to summarize the advancements in brain-inspired algorithms for intelligent robots, delving into key areas such as visual cognition, emotion-modulated decision making, musculoskeletal robotics, and motion control. Ou et al. \cite{ou2022overview} provided an introduction to brain-like computing models and chips, their evolution history, common application scenarios, and future prospects. Hassabis et al.~\cite{hassabis2017neuroscience} explored the historical connections between AI and neuroscience, and examined recent advancements in AI that are inspired by research into neural computing in both humans and other animals. In \cite{macpherson2021natural}, the authors demonstrated how machine learning and neural networks have transformed the fields of animal behavior and neuroimaging research. As for brain-inspired learning in ANNs, the biological basis and algorithm introduction can be found in~\cite{schmidgall2024brain}. This comprehensive review mostly focused on introducing how to learn from the physical structure of human brain. None of them noticed and reviewed AI models that were inspired by human behaviors and learning mechanism. Moreover, they have not comprehensively discussed which part of human brain and nervous system AI can learn to design models.

In this review, we mainly answer the following questions: What is BIAI? What's the difference between BIAI and general AI? What advantages can BIAI bring to us? Which perspectives of human brain can we learn to design AI models? What kind of BIAI models have been used in the real world? Which research areas could be further promoted by introducing BIAI? What challenges are researchers facing when integrating neuroscience with AI models? What kind of gaps are existing in the current BIAI technologies and are there any works can be done in the future? By answering these questions, we hope that researchers could increase their understanding of BIAI systems and enhance their ability to design more suitable BIAI algorithms for different applications.

\begin{figure*}
	\centering
	\includegraphics[width=1\textwidth]{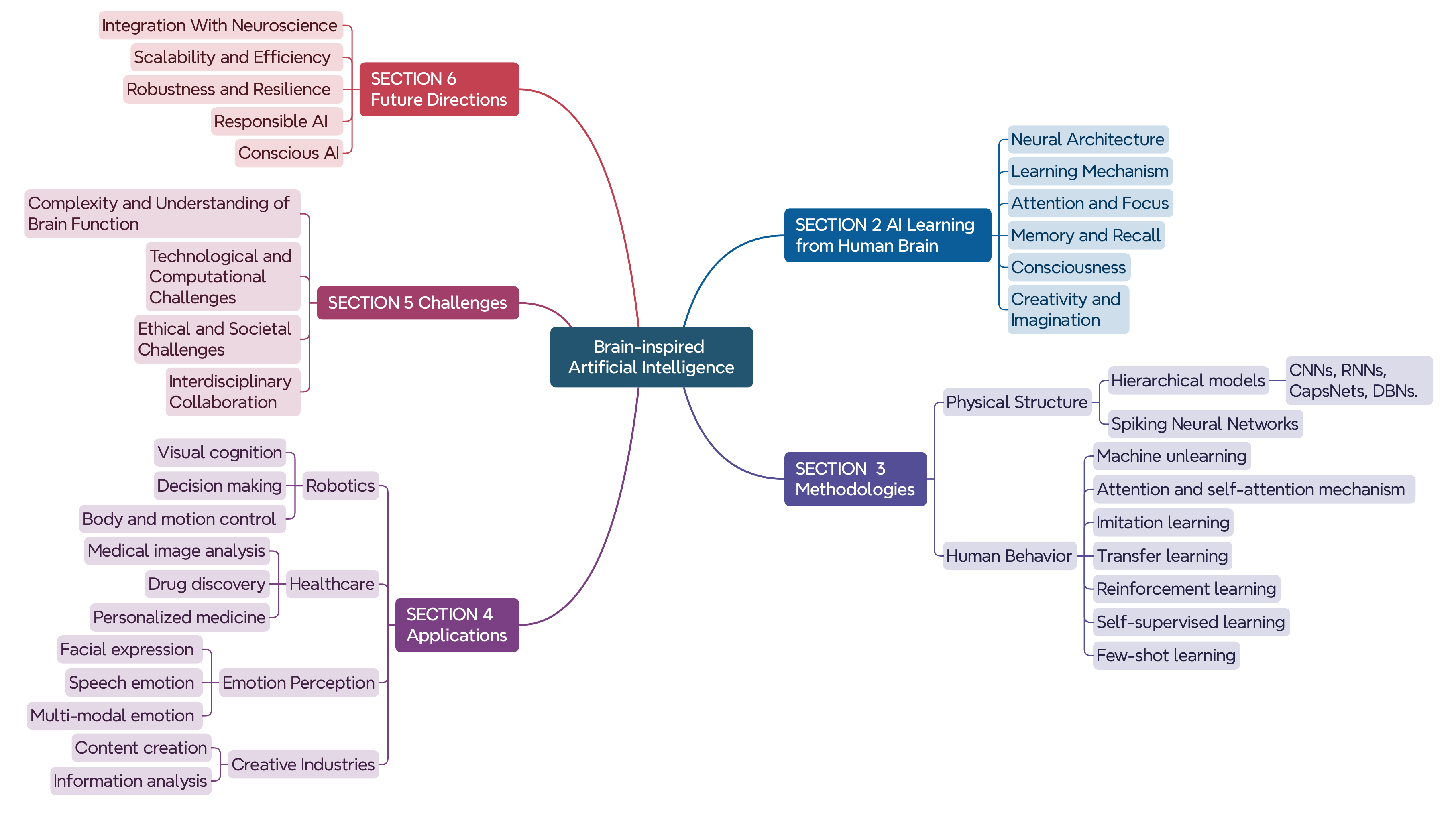}
	\caption{Structure of this comprehensive review.}\label{overview}
\end{figure*}

\subsection{Contributions}
The coverage of this paper is shown in Fig. \ref{overview}. Our main contributions are summarized as follows:

\begin{itemize}
		
		\item We introduce knowledge and insights from neuroscience and human behavior research, highlighting how AI can learn from the neural architecture, learning mechanisms, attention and focus, memory and recall, cognitive process, and creativity and imagination observed in the human brain. 
		\item We categorize BIAI studies into two main types: physical structure-inspired models and human behavior-inspired models, providing a framework for understanding different approaches in the field.	
		\item We explore diverse applications of BIAI models, including their use in robotics, healthcare, emotion perception, and creative content generation, showcasing the broad potential of these models across various domains.  
		\item We discuss the challenges faced in the development and implementation of BIAI, such as understanding brain functionality, integrating with neuroscience, and building efficient, robust, ethical, conscious, and interpretable models. We also outline future research directions to address these challenges.

\end{itemize}

The rest of the paper is organized as follows. 
Section \ref{sec2} summarizes the knowledge from neuroscience and human brain function that can inform AI systems. Following this, the review is structured according to the taxonomy presented in Figure \ref{overview}. Section \ref{sec3} discusses the primary categories of BIAI, namely, physical structure-inspired and human behavior-inspired models. Section \ref{sec4} explores the real-world applications of BIAI across various domains. In Section \ref{sec5}, we outline the general challenges faced by current BIAI methods. Section \ref{sec6} highlights several promising directions for future research. Finally, the review is concluded in Section \ref{sec7}.

\section{AI Learning from Human Brain}~\label{sec2}
In this section, we explore the inspiration source of AI models that they may refer to when designing algorithm structures. These inspirations are mainly acquired by learning knowledge from neuroscience and human behavior research. Detailed techniques of existing BIAI models will be introduced in the next section. From the previous introduction, it is evident that the human brain, being the most intricate and extraordinary organ in the body, offers numerous strengths for AI models to learn from. According to our tentative understandings of how human brain controls the bodily functions and processes, AI models may be inspired from: the neural architecture, learning mechanism, attention and focus, memory and recall, cognitive process, and creativity and imagination capacity of human brain. It should be noted that the brain possesses numerous mechanisms and processes, including many that humans have yet to discover, all of which hold potential for developing advanced AI models. Here, we will introduce several aspects that have been studied in neuroscience.
%
%
\subsection{Neural Architecture}
The neural architecture of the human brain provides the foundational blueprint for many AI models. At its core, the brain comprises billions of interconnected neurons. These specialized cells exchange information using electrical and chemical signals. Neurons interconnect to create complex networks that underpin perception, learning, memory, decision-making, and other cognitive functions. A notable feature of the brain's architecture is its capacity for plasticity—its ability to adapt and reorganize in response to new stimuli and experiences~\cite{national2000people}. This adaptability is crucial for how AI models learn from the brain.

Deep neural networks (DNNs) employ a hierarchical structure, stacking layers of artificial neurons to emulate the brain's neural organization~\cite{lecun2015deep}. Each layer processes and transforms information before passing it to the next layer, similar to the brain's network of neurons. Connections between artificial neurons, akin to synapses in the brain, are assigned weights that determine the influence of one neuron on another. These weights are adjusted during the learning process, much like how the brain refines its understanding through environmental interactions. DNNs receive feedback in the form of error signals or rewards, depending on the learning paradigm~\cite{rumelhart1986learning}, which helps the model adjust its parameters to minimize errors or maximize rewards. DNNs underpin many popular AI models. For example, convolutional neural networks (CNNs) are modeled after the brain's visual processing pathways~\cite{lecun1989backpropagation}, and recurrent neural networks (RNNs) draw on the brain's sequential processing capabilities~\cite{elman1990finding}.

\subsection{Learning Mechanism}
The human brain's learning mechanism is a sophisticated and adaptive system involving perception, memory formation, decision-making, and other cognitive processes. A key principle underlying learning in the brain is neural plasticity~\cite{von2017neural}. This concept describes the brain's remarkable ability to modify its neural connections and functions in response to new experiences and stimuli. Neural connections can be strengthened or weakened, new connections can be established, and existing ones can be eliminated based on activity patterns. AI models emulate this process by adjusting the weights between artificial neurons according to patterns found in training data. This adjustment process, known as training or learning, involves updating the model's parameters to reduce errors or enhance rewards~\cite{silver2016mastering}.

Besides, the brain encodes information in a hierarchical and distributed manner. Complex ideas are built upon simpler ones, with information spread across different brain regions. AI models, especially DNNs, use analogous principles of representation learning to understand data hierarchically. Lower layers of these models identify basic patterns, while higher layers interpret more abstract concepts~\cite{bengio2013representation}. Techniques such as transfer learning and unsupervised learning mimic this brain-like learning process. Transfer learning leverages knowledge acquired from one task to enhance performance on a related but distinct task. Unsupervised learning allows models to learn from data that is either unlabeled or only minimally supervised. Both techniques improve the model's generalization capabilities and reduce the dependency on extensive labeled datasets.

\subsection{Attention and Focus}
The attention and focus of the human brain are crucial cognitive mechanisms that allow us to selectively process information, allocate mental resources, and concentrate on specific tasks or stimuli while ignoring distractions~\cite{roda2006attention}. Understanding how attention works in the brain has inspired the development of attention mechanisms in AI models, especially in the realm of deep learning. The human brain can attend to multiple aspects of information simultaneously, a capability known as parallel or multi-head attention. This allows us to process complex stimuli and perform multiple tasks in parallel to some extent. AI models employ multi-head attention mechanisms to simultaneously attend to different parts or features of the input data. By splitting the attention mechanism into multiple heads, models can capture diverse aspects of the input and integrate information from multiple sources, improving robustness and performance~\cite{vaswani2017attention}.

Human attention is dynamic and adaptive, meaning it can shift rapidly in response to changing task demands, environmental cues, and internal states. AI models learn dynamic and adaptive attention mechanisms through training on diverse datasets and learning from feedback signals. Reinforcement learning techniques can be used to train models to adaptively allocate attention based on rewards or task performance, enabling them to learn optimal attention strategies over time~\cite{sutton1998reinforcement}. Inspired by the human brain's attentional focus, AI researchers have developed sophisticated attention mechanisms. These mechanisms significantly enhance the performance and interpretability of AI models across various applications~\cite{rigotti2021attention}. 

\subsection{Memory and Recall}
The human brain comprises multiple memory systems~\cite{atkinson1968human}. Sensory memory temporarily holds sensory information from the environment and acts as a buffer for stimuli received through sight, hearing, touch, smell, and taste. Short-term memory serves as a temporary storage system with limited capacity, maintaining information that is readily accessible for immediate use. Long-term memory is responsible for storing information for extended periods, potentially without limit. AI models can emulate these memory systems using various architectures and mechanisms. For example, RNNs and transformers employ recurrent connections and attention mechanisms to model short-term dependencies and long-term context in sequential data, enabling the retention and retrieval of information over time.

Recall refers to the process of accessing and retrieving information stored in memory. It can occur spontaneously or be triggered by external cues or internal associations. AI models perform recall and retrieval by accessing stored representations through inference or query mechanisms~\cite{kirkpatrick2017overcoming}. Memory-based architectures, such as memory networks and neural Turing machines, enable models to retrieve relevant information from memory based on input queries or context. By drawing inspiration from the memory and recall mechanisms of the human brain, AI researchers have developed memory-augmented architectures and learning algorithms that enhance the capabilities of AI models in storing, retrieving, and leveraging information over time. Long-Short-Term Memory (LSTM) neural networks~\cite{hochreiter1997long} are among the most commonly used memory models in AI.

\subsection{Consciousness}
While consciousness remains a subject of philosophical and scientific exploration, it is closely tied to cognitive processes like attention, memory, decision-making, and self-awareness~\cite{rumelhart1986learning}. Although AI models do not experience consciousness as humans do, they can leverage aspects of conscious processing to enhance their performance and capabilities. Consciousness supports social cognition, including the ability to understand and predict others' mental states, beliefs, and intentions. This is known as the theory of mind, which facilitates social interactions and empathy.

AI models can draw from social cognition and theory of mind to create more human-like interactions and behaviors in social contexts. Techniques like sentiment analysis and empathy modeling allow models to interpret and respond to users' emotional states and intentions, thereby improving human-AI interactions~\cite{vodrahalli2022humans}. While AI does not possess consciousness per se, learning from the principles of conscious processing can lead to more sophisticated and effective models and algorithms. By incorporating these cognitive processes, researchers aim to develop AI systems with increasingly advanced behaviors and interactions.

\subsection{Creativity and Imagination}
The creativity and imagination of the human brain are remarkable cognitive abilities that enable us to generate novel ideas, insights, and solutions, as well as to envision hypothetical scenarios and possibilities~\cite{turner2014origin}. Creativity often involves the ability to combine existing concepts, ideas, or elements in novel and unexpected ways. The human brain achieves this by flexibly manipulating and recombining mental representations. AI models learn from flexible representation learning techniques that enable them to generate new and diverse outputs. Variational autoencoders (VAEs)~\cite{kingma2013auto}, generative adversarial networks (GANs)~\cite{goodfellow2020generative}, and transformers~\cite{vaswani2017attention,shehzad2024graph} with self-attention mechanisms allow models to generate realistic and novel content by learning rich and structured representations of data.

Creativity often involves emotional and aesthetic dimensions, such as appreciation of beauty, novelty, and emotional resonance. Emotions play a crucial role in guiding creative expression and evaluation. AI models learn from emotional and aesthetic dimensions of creativity through sentiment analysis, style transfer, and affective computing techniques. Models can generate content with desired emotional qualities, adapt to user preferences, and create aesthetically pleasing outputs~\cite{xu2024beyond}. Furthermore, creativity often involves making connections between seemingly unrelated domains or concepts, drawing analogies, and transferring knowledge from one context to another~\cite{zhang2022expressing}. AI models learn from analogical reasoning and transfer learning techniques that enable them to generalize knowledge and skills across different domains. Transfer learning~\cite{pan2009survey}, few-shot learning~\cite{wang2020generalizing}, and meta-learning algorithms~\cite{smith2009cross} allow models to transfer knowledge from related tasks or domains to new, unseen tasks, facilitating creativity and adaptability.

\section{Brain-inspired AI Models}~\label{sec3}
In this section, we categorize existing BIAI techniques based on the specific brain-inspired mechanisms they incorporate. By drawing insights from neuroscience, we can apply this knowledge to computational modeling, which involves creating and implementing algorithms and systems that emulate the brain's structure and function. This process can be divided into two categories: models and algorithms inspired by the brain's physical architecture, and those inspired by human behavior. The introduction order of these techniques will mainly follow the order shown in Fig.~\ref{overview}.
\subsection{Physical structure-inspired AI Models}
Several prominent AI models seek to emulate the human brain's architecture and processes, leveraging neuroscience to develop more biologically plausible and powerful systems. Here are a few notable examples. We will only list the most representative models and their concepts in this section to help understand the theoretical basis and internal mechanism of these models.
\subsubsection{Hierarchical Models}
A representative mechanism of the PS-inspired model is to learn from the information processing meachanism of human brain. The human brain processes information in a hierarchical manner, with sensory inputs processed at lower levels and abstract concepts at higher levels. Learning from this process, DNNs where entities are organized into levels or layers, each layer representing different levels of abstraction or processing are designed. Representative hierarchical models include CNNs, Capsule Networks (CapsNets), RNNs, Echo State Networks (ESNs), and Deep Belief Networks (DBNs). Considering that these models are the most basic models that have been used in different algorithms, we only introduce the basic concepts and how they are inspired by human brain. As for the characteristics, realization, equations, and real application scenarios of these models, we refer readers to~\cite{zhao2024review,lalapura2021recurrent,wang2020survey} for more details.

\textbf{CNNs} are a specialized type of ANNs designed for processing structured grid data. The concept was inspired by the work of biologists Hubel and Wiesel, who conducted pioneering research on the visual cortex of the cat brain~\cite{jogin2018feature}. They discovered the phenomenon known as receptive field mechanisms: neurons in the primary visual cortex respond to specific features in the visual environment. Hubel and Wiesel identified two types of cells: simple cells respond strongly to specific spatial positions and orientations, while complex cells have larger receptive fields and remain responsive even when the position of the features shifts slightly. 

When we view an image, our attention is initially drawn to the most prominent and informative local features. For instance, in a picture of a pet dog, our eyes typically focus on the dog first, noticing its face, legs, and other distinct parts. Based on our experience, we recognize that it is a picture of a dog. CNNs emulate this principle of the human visual system. Initially, a CNN identifies low-level features by detecting edges and curves within the image. Through a series of convolutional layers, these low-level features are gradually combined into more complex, high-level features~\cite{krizhevsky2017imagenet}. Since high-level features are composed of multiple low-level feature convolutions, they encapsulate more comprehensive information from the original image.

CNNs use convolutional layers to apply filters to input data, extracting hierarchical features like edges, textures, and shapes. This enables CNNs to effectively capture spatial dependencies and patterns within the data. Following these layers, pooling layers reduce the spatial dimensions, which helps manage computational complexity and mitigate overfitting. The integration of these layers, along with fully connected layers at the end, enables CNNs to excel in tasks such as object detection, segmentation, and image classification, establishing them as a fundamental component of modern computer vision applications~\cite{li2021survey}. However, CNNs also face challenges~\cite{krichen2023convolutional}, including the need for large labeled datasets to achieve high performance, high computational costs, and the potential loss of valuable spatial information due to pooling layers. Additionally, CNNs can struggle with understanding spatial hierarchies and complex part-whole relationships, which has led to the development of alternative architectures like capsule networks to address these limitations.

%
\textbf{CapsNets}, proposed by Geoffrey Hinton and his team in 2017~\cite{sabour2017dynamic}, aim to challenge the traditional CNN architecture by addressing a critical flaw in CNNs: the use of pooling layers. Hinton argues that pooling, particularly max pooling, is a significant mistake despite its widespread success. The max pooling process discards valuable spatial information by only passing the most active neurons to the next layer, leading to a loss of crucial details between layers. CapsNets aim to preserve this spatial hierarchy, offering a more nuanced understanding of the relationship between parts and wholes in an image, thereby potentially improving performance on tasks requiring precise spatial information.

CapsNets use capsules, which are neuron groups encoding both the probability of an entity's presence and its properties, thereby preserving more detailed spatial information. This leads to better generalization and robustness, especially in tasks involving viewpoint variations and part-whole relationships. However, CapsNets also face challenges~\cite{dombetzki2018overview}, such as higher computational complexity and the need for more sophisticated training algorithms, which can make them less efficient and harder to implement compared to CNNs. Additionally, the field is still developing, and practical applications and optimizations are ongoing areas of research.

CapsNets can preserve spatial hierarchies and relationships within data, making them effective in various applications. One notable application domain is medical imaging~\cite{yadav2024te}, where CapsNets have demonstrated improved accuracy in tasks such as tumor detection and organ segmentation by maintaining spatial integrity of anatomical structures. In computer vision, CapsNets are used for image classification and object recognition, providing better robustness to affine transformations compared to traditional CNNs~\cite{pawan2022capsule}. Additionally, they are being explored in NLP tasks like sentiment analysis and entity recognition, benefiting from their capability of capturing complex relationships within sequences of text. These applications highlight the potential of CapsNets to enhance the performance and reliability of machine learning models in diverse and complex domains.

\textbf{RNNs} are built to process sequential data by retaining internal memory states. They are inspired by the human brain's ability to process sequences of information and retain memory over time~\cite{lipton2015critical}. Just as our brain remembers previous experiences and uses them to understand current events, RNNs use their internal memory to process and remember information from previous steps in a sequence. For example, consider how we understand a sentence. When reading the sentence "the cat sat on the mat," our brain doesn't start from scratch with each word. Instead, it retains the context provided by earlier words to make sense of the entire sentence. We remember "the cat" as we read "sat on the mat", allowing us to understand the sentence's meaning in context.

Similarly, RNNs manage sequences by maintaining a hidden state that stores information from earlier steps. With each new step, the RNN updates this hidden state by incorporating both the current input and the hidden state from the previous step. This mechanism allows RNNs to retain context and make informed predictions or decisions based on the entire sequence of data. For instance, RNNs can be used for language translation. When translating a sentence from English to French, an RNN processes the English words one by one, retaining the context of the entire sentence. This context helps RNN generate the correct translation in French, even for complex sentence structures where the relationship between words must be understood in sequence. LSTM~\cite{hochreiter1997long} and Gated Recurrent Unit (GRU)~\cite{chung2014empirical} are advanced RNN architectures developed to address the vanishing gradient problem. They enable the effective capture and utilization of long-range dependencies in sequential data.

RNNs have found numerous real-world applications due to their ability to process and analyze sequential data. In the field of NLP, for instance, RNNs drive advances in machine translation, sentiment analysis, language modeling, and many other applications~\cite{min2023recent}, enabling more accurate, context-aware text generation and interpretation. In the speech recognition domain, RNNs are used to convert spoken language into written text, enhancing the functionality of virtual assistants~\cite{saon2021advancing}. In finance, RNNs help in predicting stock prices and analyzing market trends by processing time series data~\cite{rundo2019machine}. Additionally, RNNs are employed in healthcare for early diagnosis by analyzing patient history~\cite{pham2017predicting} and in video analysis for understanding and predicting sequences of frames~\cite{fan2019cubic}, making them invaluable in diverse fields that require sequence prediction and temporal pattern recognition.
%
%

\textbf{DBNs} are generative models composed of stacked Restricted Boltzmann Machines (RBMs). These models learn layer-by-layer through unsupervised techniques like contrastive divergence~\cite{hinton2006fast}. Information from sensory inputs is processed through multiple stages, with each stage extracting more abstract and complex features from the input. Similar to human brain, DBNs process data through multiple layers, with each layer learning increasingly complex representations. DBNs are used for tasks such as unsupervised feature learning, dimensionality reduction, and collaborative filtering.

Thanks to their capacity to learn complex representations from big data, DBNs have been utilized in numerous real-world applications. In speech recognition, for example, DBNs have been used to enhance transcription accuracy by learning hierarchical audio features~\cite{mohamed2011acoustic}. In image recognition, they are used to enhance object detection and classification tasks by capturing intricate patterns and textures in images~\cite{abdel2016breast}. In NLP, DBNs help in understanding and generating human language, improving the performance of sentiment analysis and other tasks~\cite{sarikaya2014application}. Additionally, DBNs are applied in recommendation systems, where they analyze user behaviors and preferences to suggest relevant content or products~\cite{wang2014improving}. Their ability to model deep, layered representations makes DBNs valuable in any domain requiring sophisticated data interpretation and prediction.

These hierarchical models have many advantages~\cite{liu2017survey}. Firstly, DBNs can autonomously learn rich hierarchical features from unlabeled data. This eliminates the need for manual feature extraction. Secondly, by extracting abstract features layer by layer, DBNs effectively capture the deep structure of data, which is beneficial for handling complex, high-dimensional problems. Thirdly, the pretraining phase helps in escaping local optima and provides a good initialization for subsequent fine-tuning. However, implementing hierarchical models can be challenging due to the complexity of designing and training multi-level architectures. DBNs consist of multiple layers of RBMs, and the training process involves multiple iterations of contrastive divergence, which is computationally intensive and has a slow convergence rate. Additionally, deep networks are prone to overfitting, especially when the amount of data is limited. It is necessary to employ strategies such as regularization and early stopping to prevent overfitting. As a deep generative model, the decision-making process of a DBN is difficult to understand and lacks intuitive interpretability.

\subsubsection{SNNs}
Spiking neural networks depart from traditional neural networks by modeling neurons as spiking units, aligning more closely with the brain's computational mechanisms~\cite{tavanaei2019deep}. They belong to the third generation of neural network models, achieving a more advanced level of biological neural simulation. 

SNNs use discrete events called spikes to represent neuron activations. These spikes are initiated when a neuron's membrane potential attains a critical threshold, closely replicating the mechanism of action potentials in biological neurons. SNNs can be distinguished from traditional DNNs in four key aspects~\cite{wang2020supervised}: 1. SNNs employ temporal encoding, representing information as spike trains, while DNNs use rate encoding, converting data into scalar values. This temporal encoding grants SNNs superior capabilities for handling complex temporal or spatiotemporal data. 2. SNNs are built upon spiking neurons modeled by differential equations, contrasting with the artificial neurons and activation functions used in DNNs. Moreover, SNN simulations often leverage clock-driven or event-driven approaches, differing from DNNs' step-by-step method. 3. SNNs adopt spike-timing-dependent plasticity (STDP) for synaptic learning, departing from DNNs' reliance on the Hebbian rule. Training methodologies also vary, with SNNs exploring diverse techniques compared to DNNs' predominant use of loss function derivatives. 4. SNNs excel at rapid, parallel processing due to their discrete spike-based computations, offering potential energy efficiency advantages over DNNs' analog-like operations~\cite{farsa2019low}.


The inspiration for SNNs comes from the way the human brain processes information~\cite{ghosh2009spiking,yao2023attention}. In the brain, neurons communicate via electrical impulses or spikes, which encode information through the timing and frequency of these spikes. This spiking mechanism allows for highly efficient and precise information processing. SNNs attempt to replicate this by using similar spiking mechanisms to transmit information between artificial neurons. This leads to potentially more efficient computing, especially in tasks involving temporal patterns and sequences, such as speech and motion recognition. By capturing the temporal dynamics of neural processing, SNNs aim to achieve more powerful and efficient models for complex tasks~\cite{zheng2021going}. SNNs can also leverage spiking behavior for real-time processing, making them suitable for applications in robotics and neuromorphic computing~\cite{lele2021end}. However, SNNs face several challenges, including the complexity of designing and training models as spiking neurons are non-differentiable by nature, which complicates the use of traditional gradient-based optimization methods~\cite{che2022differentiable}. Additionally, there is a lack of mature software and hardware infrastructure compared to more established neural network models, making practical implementation and scaling of SNNs more difficult~\cite{zhang2020low}. Despite these challenges, ongoing research aims to harness the potential of SNNs for advanced AI applications.

\subsection{Human behavior-inspired AI Models}
There are also many AI models inspired by human behaviors and cognitive processes in several ways. The learning mechanisms of these models usually leverage insights from cognitive psychology, neuroscience, and behavioral science to create more effective and intelligent systems. Instead of introducing the specific models, we only summarize and classify the most popular learning mechanisms of these models in recent years. In each learning paradigm, we select a representative situation to concrete and better understand the internal operating mechanisms of these models. 
\subsubsection{Machine Unlearning}
\begin{figure*}[htb]
	\centering
	\includegraphics[width=0.9\textwidth]{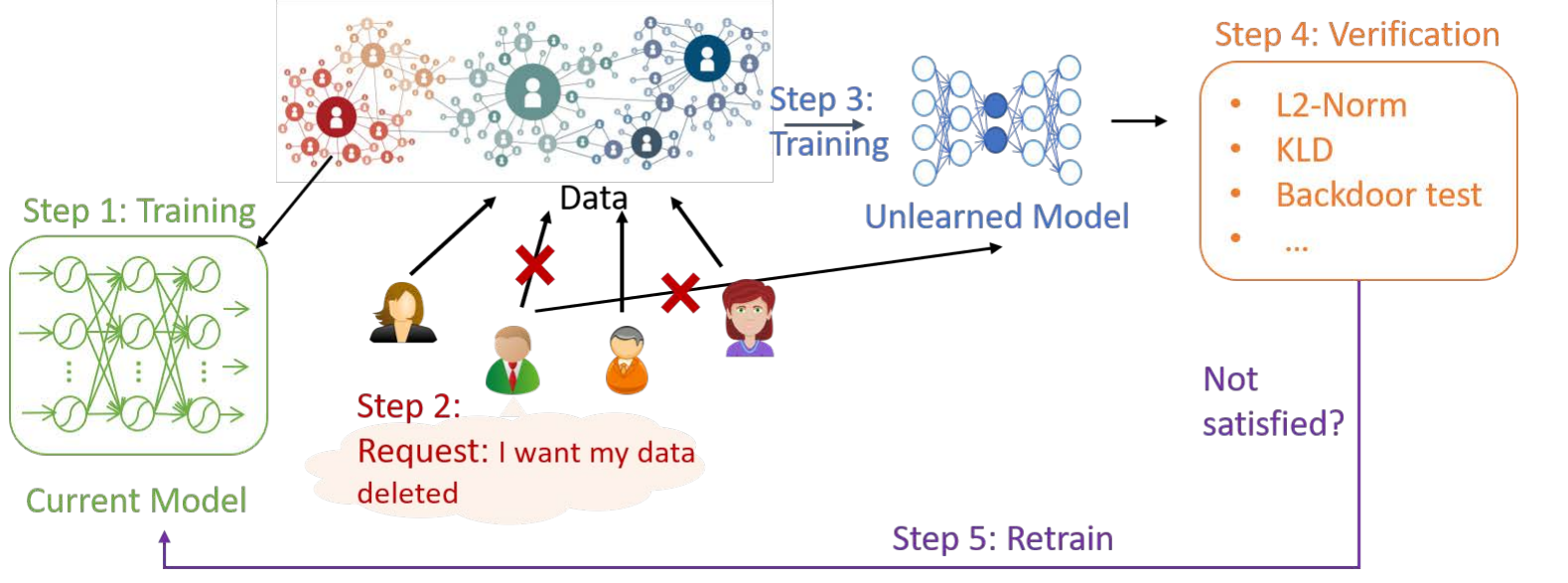}\\
	\caption{The process of machine unlearning. When users want their data deletion, they should be deleted from the database and it needs to remove their contribution to the already trained model.}\label{unlearning}
\end{figure*}

It involves selectively removing the influence of specific data points from a trained machine learning model. It can be used to protect users’ privacy when they want their data forgotten by a trained machine learning model. Instead of simply erasing their data from database, the deletion needs to eliminate the contribution of the erased training data from the already trained model, which process was first defined as machine unlearning~\cite{wang2024machine}. To better understand how unlearning mechanisms work, we first introduce the unlearning problem and process following the machine unlearning framework demonstrated in Fig~\ref{unlearning}. Assuming that a model $M$ is trained on the dataset $D$ with algorithm $\mathcal{A}$, this could be denoted as
\begin{equation}
M=\mathcal{A}(D),
\end{equation}
where $D\in \mathcal{Z}$. Here, $\mathcal{Z}$ is the space of data items and model $M$ is in a hypothesis space $\mathcal{H}$. If a user wants to remove his data $D_e$ from the trained model, then the server removes the contribution of $D_e$ using a machine unlearning algorithm $\mathcal{U}$, which is denoted as
\begin{equation}
M_{D\setminus D_e}=\mathcal{U}(D,D_e,\mathcal{A}(D)).
\end{equation}
After providing an unlearning request, the unlearned model $\mathcal{U}(D,D_e,\mathcal{A}(D))$ is expected to be same as the retrained model $\mathcal{A}(D\setminus D_e)$. The evaluation distance metrics in the verification process are diverse, including 
\begin{equation}
L_2-norm = \lVert \theta_{\mathcal{A}(D\setminus D_e)}-\theta_{\mathcal{U}(D,D_e,\mathcal{A}(D))} \rVert,
\end{equation}
and
\begin{equation}
KL-Divergence = KL[\theta_{\mathcal{A}(D\setminus D_e)} \lVert \theta_{\mathcal{U}(D,D_e,\mathcal{A}(D))}],
\end{equation}
where $\theta_{\mathcal{A}(D\setminus D_e)}$ is the model parameters of retraining from the scratch and $\theta_{\mathcal{U}(D,D_e,\mathcal{A}(D))}$ is the parameters of unlearning algorithms $\mathcal{U}(\cdot)$.

This technique is crucial for addressing privacy concerns, correcting errors, and maintaining model relevance as data evolves. By effectively "forgetting" certain data, models can comply with privacy regulations, such as General Data Protection Regulation (GDPR)~\cite{mantelero2013eu}. Machine unlearning enhances model efficiency and fairness by ensuring that outdated, incorrect, or biased data does not negatively impact predictions~\cite{tu2023deep}, while avoiding the computational cost and time required for complete model retraining.

The theoretical foundation of machine unlearning is inspired by the brain's ability to selectively forget and rewire neural connections, a process known as synaptic pruning~\cite{wang2024machine}. This mechanism enables the brain to maintain cognitive efficiency by removing less useful or redundant information, thereby optimizing memory storage and retrieval. Similarly, machine unlearning seeks to enhance machine learning models by selectively eliminating the impact of particular data points, ensuring that the model remains efficient and accurate without being burdened by obsolete or incorrect information. This parallel to human cognitive processes highlights the importance of adaptability and memory management in both biological and artificial systems.

Most machine unlearning models are designed to delete regularly structured data, aiming primarily to reduce computation and storage costs while minimizing performance degradation. Recently, some researchers have also focused on achieving unlearning in scenarios involving irregular data, such as graph unlearning~\cite{chen2022graph}. Graph-structured data are more complex than standard structured data because they encompass not only the feature information of individual nodes but also the connectivity information inside edges between different nodes~\cite{ren2023graph}. The first step in general process of graph unlearning is to cut off some connecting edges to split the graph into several sub-graphs. Then, the constituent graph models could be trained on these subgraphs respectively, and ensemble them for final prediction. As for the knowledge unlearning of a specific node, a method to efficiently remove node features by projecting model parameters onto an irrelevant subspace has been proposed~\cite{cong2023efficiently}. In this model, challenges caused by node dependency could be solved, and the deleted node features are guaranteed to be unlearned from the pre-trained model.

\subsubsection{Attention and Self-Attention Mechanism}

In AI, the attention mechanism enables machine learning models to focus on relevant input data, leading to improved performance. Inspired by human cognitive processes~\cite{posner1990attention}, particularly selective attention in psychology, attention mechanisms allow models to dynamically prioritize different parts of input data. By assigning varying degrees of attention, typically through learned weights~\cite{niu2021review}, the mechanism enables models to prioritize certain features or contextually relevant information while filtering out noise or irrelevant details. This adaptive focus not only improves the model's performance in terms of, e.g., accuracy and efficiency in many tasks like NLP, image recognition, and sequence prediction but also enhances its interpretability by revealing which parts of the input are most influential in generating predictions or outputs~\cite{hsu2019multivariate}.

Despite that models inspired by the attention mechanism~\cite{vaswani2017attention,bahdanau2015neural} have been proposed and successfully applied into different scenarios, the mechanisms underlying consciousness are still not well understood. Creating AI that can mimic or exhibit conscious-like states is a profound challenge. For example, emotions are vital to decision-making and learning. However, the neural basis of emotions and their integration with cognitive processes is not fully understood. Grabbing knowledge of this process can help complete tasks like sentiment analysis, and emotion perception in robotics~\cite{destephe2013perception}.

The well-known Transformer is a powerful neural network architecture introduced in 2017~\cite{vaswani2017attention}, designed to handle sequential data for tasks like language translation and text generation. Generative Pre-trained Transformers (GPTs) process data differently than RNNs, using self-attention to weigh input elements independently. This enables them to capture long-range dependencies more effectively. Their architecture, composed of encoder and decoder layers with self-attention and feed-forward networks, makes them versatile for NLP tasks~\cite{zhu2021long}. While traditional Transformers are widely used in NLP and other sequential data tasks, Graph Transformers~\cite{shehzad2024graph} extend these concepts to handle the complexities of graphs, where data is represented as nodes and edges instead of sequences.

Large Language Models (LLMs) have recently captured significant attention for their robust text generation and comprehension abilities. By fine-tuning these models with specific instructions, they can be better satisfied with user intent, demonstrating strong interactive capabilities as well as the potential to boost productivity as intelligent assistants. Due to the limitations of LLMs suffering from processing other common modalities like images, speech, and videos, a new class of Vision-Language Models (VLMs) has been developed. These models enhance LLMs by enabling them to process and understand visual information, demonstrating potential for solving vision-centric problems. To better understand these models, we summarize some major GPTs developed in recent years in Table\ref{gpt}.

\begin{table}[htbp]
	\caption{\label{gpt}Examples of popular GPTs}
	\begin{tabular}{|p{2.5cm}<{\centering}|p{5cm}<{\centering}|p{2cm}<{\centering}|p{1.5cm}<{\centering}|p{2cm}<{\centering}|}
		\toprule
		\textbf{Name} & \textbf{Developer} &\textbf{Modal}&\textbf{Year} & \textbf{Ref.}\\
		\midrule
		GPT-4V & OpenAI  & Multi &2024 & \cite{achiam2023gpt}\\	\hline
		Gemini & Gemini Team, Google&  Multi&2023& \cite{team2023gemini}\\	\hline
		LLaMA & Meta AI & Single&2023& \cite{touvron2023llama}\\	\hline
		Claude-3 & Anthropic & Multi& 2024& \cite{anthropic2024claude}\\	\hline
		Mistral 7B & Mistral AI& Single &2023& \cite{jiang2023mistral}\\	\hline		
		Grok-1.5V & xAI & Multi&2024& \cite{xai2024grok}\\	\hline	
		MM1 & Apple& Multi&2024& \cite{mckinzie2024mm1} \\	\hline	
		Step-1V & StepFun Research Team &Multi&2024& \cite{step2024step} \\	\hline	
	    InternVL 1.5 & Shanghai AI Laboratory & Multi &2024&  \cite{chen2024internvl,chen2024far}\\	\hline	
		Qwen-VL & Qwen Team, Alibaba Group & Multi&2023& \cite{bai2023qwen}\\	\hline	
		EMU & Beijing Academy of AI & Multi &2024& \cite{sun2023emu,sun2024generative}\\	\hline	
		GLM-4V & Team GLM, Zhipu AI&Single&2024 & \cite{team2024chatglm} \\	\hline

		\bottomrule
	\end{tabular}
\end{table}

As mentioned above, attention mechanisms enhance model performance by focusing on relevant input parts, improving long-range dependency handling. Particularly in NLP, they prioritize certain words, boosting tasks like translation and generation. Additionally, attention offers insights into model decision-making. However, attention mechanisms also present several challenges~\cite{han2022survey}. Self-attention computations can be computationally expensive, especially for long sequences due to their quadratic complexity. Furthermore, designing effective attention mechanisms requires careful consideration of hyperparameters and architectural choices, which can be complex and task-specific. Despite these challenges, the attention mechanism remains a powerful tool in AI, driving advancements in various fields and continually inspiring new research directions.

\subsubsection{Imitation Learning}
It is a machine learning approach where agents learn tasks by copying expert behavior. This is often more efficient than trial-and-error, as it leverages expert knowledge to avoid exploring countless potential action sequences~\cite{calinon2009robot}. This approach tends to leverage demonstrations (pairs of input and output) provided by the expert to train the agent, enabling it to acquire complex skills more efficiently. By observing and replicating the expert's actions, the agent can generalize from these examples to handle similar situations in the future. Imitation learning is particularly useful in environments where explicit programming of all possible actions is impractical. Its effectiveness has been demonstrated across various domains, including robotics, video games, object manipulation, autonomous driving, among others.

Imitation learning is inspired by humans' ability to learn by observing and copying others. Humans and animals learn by watching others, a process known as observational learning. Neuroscience studies have shown that observing others can lead to learning without direct experience. Imitation learning algorithms allow agents to learn from demonstrations without needing explicit programming or direct environment interaction~\cite{hussein2017imitation}. Neuroscientific studies suggest that mirror neurons, a type of neuron found in areas of the brain responsible for motor control and understanding intentions~\cite{gallese2009motor}, play a crucial role in this process. These neurons are triggered not only when an individual carries out an action but also when they witness another person executing the same action, thus facilitating learning through imitation. Neural networks can simulate this process by learning patterns from expert demonstrations to mimic their behavior. 

Behavioral Cloning (BC) and Inverse Reinforcement Learning (IRL) are widely used techniques in imitation learning, but both face a common issue: they require substantial high-quality demonstration data. To overcome this, Generative Adversarial Imitation Learning (GAIL)~\cite{ho2016generative} was introduced in 2016, combining GANs~\cite{goodfellow2014generative} and maximum entropy IRL. GAIL uses a generator and a discriminator. The generator aims to produce data indistinguishable from expert demonstrations, while the discriminator differentiates between real and generated data. Compared to other previous algorithms, GAIL is more efficient in utilizing expert data and eliminates the need for ongoing expert interaction during training.

\subsubsection{Transfer Learning}
As a popular machine learning technique, transfer learning applies knowledge from one task to improve performance on a different but related task. For instance, a child's surprise at a floating toy reveals the transfer of learned knowledge about gravity, even without explicit teaching~\cite{lecun2018power}.

According to the generalization theory of transfer, the capacity to transfer knowledge stems from the ability to generalize experience~\cite{zhuang2020comprehensive}. In other words, when a person applies their past experiences to new situations, they enable the possibility of transferring knowledge from one context to another. This theory posits that a connection between two learning activities is necessary for transfer to occur. For instance, a violinist often learns piano faster due to shared musical fundamentals. 


Transfer learning applies knowledge gained in one area to improve learning in a related field, mirroring human ability to apply learned skills to new situations.
A prominent example of transfer learning is the fine-tuned BERT (Bidirectional Encoder Representations from Transformers)~\cite{kenton2019bert}. BERT is pre-trained on a vast text corpus to grasp language contexts using a transformer architecture that processes text bidirectionally. Adapting it to specific tasks with minimal data significantly boosts performance. This highlights transfer learning's potential to enhance model capabilities, especially in data-scarce domains~\cite{pan2009survey}. However, it is worth noting that knowledge transfer does not always lead to positive outcomes for new tasks; if the domains share little in common, the transfer may be ineffective.

\subsubsection{Reinforcement Learning (RL)} It is a popular machine learning approach in which agents learn to make decisions through interactions with an environment~\cite{wang2022deep}. The goal of the agent is to maximize rewards by selecting optimal actions. Learning occurs by means of trial-and-error, in which the agent receives feedback in the form of rewards or penalties. RL is particularly effective for problems where the correct actions are not immediately clear and must be discovered through a balance of exploration and exploitation.

RL draws substantial inspiration from behavioral psychology, particularly the principles of operant conditioning \cite{barto2013intrinsic}. In operant conditioning, behaviors are influenced by their consequences: actions leading to rewards (positive reinforcement) are more likely to be repeated, whereas those followed by punishments (negative reinforcement) are less likely to occur again. RL mirrors this framework by having an agent interact with an environment, perform actions, and receive feedback in the form of rewards or penalties. This feedback loop enables the agent to learn optimal behaviors to maximize cumulative rewards. Furthermore, RL incorporates the concepts of exploration (trying new actions to observe their effects) and exploitation (choosing known actions that yield high rewards), akin to how humans balance curiosity with the use of established strategies to learn from their environment \cite{rivest2004brain}. These parallels illustrate how RL algorithms emulate human learning processes to develop intelligent, adaptive systems.

Deep Reinforcement Learning (DRL) integrates RL with DNNs to facilitate agents' learning and decision-making in complex, high-dimensional environments. DRL takes advantage of the deep networks' powerful ability to learn representations from raw sensory inputs, e.g., images or audio, to extract relevant features that guide the agent's decisions. A notable example of DRL is Deep Q-Networks (DQN)~\cite{mnih2015human}, which uses a CNN to estimate the value of actions in different states. The network is trained to minimize the difference between predicted and actual rewards.

By combining the strengths of deep learning and RL, DRL excels in handling high-dimensional input spaces, such as images or raw sensory data, by leveraging DNNs to extract and learn relevant features automatically. This capability allows DRL agents to make decisions directly from complex and unstructured data without requiring extensive feature engineering. Additionally, DRL can learn sophisticated policies and strategies in environments with complex dynamics and long-term dependencies, making it suitable for tasks like game playing, robotics, autonomous driving, and resource management. The ability to learn from trial and error and continuously improve through interacting with the environment enables RL to solve problems that are challenging for traditional machine learning approaches.


\subsubsection{Self-supervised Learning (SSL)}
Supervised deep learning models generally require vast amounts of labeled data, which can be costly and time-consuming to obtain. To address this, SSL extracts meaningful features from unlabeled data without human intervention~\cite{gui2024survey}. This method typically involves tasks like predicting missing segments of input data, generating data transformations, or distinguishing between modified and original data samples. By creating pretext tasks—artificial labels generated from the data itself, such as predicting the next frame in a video, completing missing parts of an image, or determining the temporal order of events, the model trains itself to understand complex patterns and structures. This method leverages the inherent information in the data to build robust and transferable features, reducing the reliance on expensive and time-consuming human annotations. Deep SSL has significantly advanced fields like speech recognition, NLP, and computer vision by achieving remarkable success with unlabeled data.

SSL is inspired by the human brain's ability to learn from its environment without explicit supervision~\cite{liu2021self,liu2022graph}. The human brain continuously processes vast amounts of sensory information and extracts patterns, relationships, and structures through experiences and observations. Similarly, SSL algorithms generate supervisory signals from the data itself by creating pretext tasks, such as the prediction of missing values of an input or the relationship between different parts~\cite{ren2022eagle}. This approach allows models to learn meaningful representations and features from unlabeled data, much like how humans learn to understand and interpret the world around them by making sense of incomplete information and drawing connections between different experiences.

SSL offers several advantages, including the ability to leverage vast amounts of unlabeled data and reduce reliance on expensive labeled datasets. SSL creates surrogate tasks like masked word prediction or jigsaw puzzles to learn valuable data representations, which can then be fine-tuned for specific applications~\cite{xie2022self,yu2023self}. However, SSL also faces open issues, such as the challenge of designing effective pretext tasks that lead to useful representations, the difficulty of scaling methods to complex tasks and diverse domains, and the risk of learned representations being biased or not generalizing well to all applications. Additionally, balancing the trade-offs between computational resources for training and the performance gains achieved remains an ongoing area of research~\cite{zhang2024self}.

\subsubsection{Few-shot learning (FSL)}
Traditional machine learning and deep learning models struggle with extremely limited data. When models are trained on only a few hundred or fewer examples, they often overfit, memorizing the training data rather than learning general patterns, which might potentially result in poor performance on new data. However, there is a widespread demand in the industry for training models based on small samples, such as single-user face and voice recognition, recommendation cold-start, fraud detection, and other scenarios where sample sizes are small or data collection costs are high. FSL addresses the challenge of learning from limited data by incorporating additional information. This includes unlabeled data, other datasets, or prior knowledge to enhance learning from scarce labeled samples~\cite{song2023comprehensive}.
 
FSL refers to the capability of a model to learn from extremely limited data~\cite{wang2020generalizing}. Unlike traditional machine learning, which often requires vast amounts of labeled examples, FSL aims to achieve high performance with only a handful of training instances. This approach is especially valuable in situations where acquiring supervised examples is challenging or impossible due to privacy, safety, or ethical concerns~\cite{song2023comprehensive,antonelli2022few}. Drug discovery, for instance, often faces challenges due to the limited availability of real biological data on clinical candidates~\cite{altae2017low}. Identifying new molecules for drug development requires learning from scarce data due to factors like toxicity, low activity, and solubility. This necessitates efficient learning from minimal examples.

FSL in machine learning takes inspiration from neuroscience, especially the brain's exceptional capability to learn new concepts with minimal examples~\cite{bontonou2021few}. Humans can quickly learn new tasks by leveraging their previous knowledge and experiences. They generalize from just a few instances due to mechanisms like synaptic plasticity, which allows neurons to adjust their connections based on experience~\cite{soltoggio2018born}. Similarly, FSL algorithms employ techniques such as meta-learning~\cite{li2020few}. By training on diverse tasks, models develop a general learning ability that can be quickly adapted to new, data-scarce scenarios~\cite{li2021concise}. This approach mirrors the brain's efficiency in leveraging prior knowledge and context to understand and learn from new situations rapidly.

FSL offers significant advantages by enabling models to learn effectively from limited data, mimicking human learning. This is crucial in data-scarce environments, reducing reliance on extensive labeled datasets and accelerating training. By quickly adapting to new tasks with minimal data, FSL enhances model adaptability~\cite{baik2023learning}. However, several challenges remain, including achieving robust generalization across diverse and complex tasks, sensitivity to the quality and representativeness of the provided examples, and the need for more efficient algorithms to manage the high computational demands often associated with these models~\cite{song2023comprehensive}. Overcoming these challenges is essential for further advancing the practical applications of FSL.

\section{Applications}~\label{sec4}
\begin{figure*}[htb]
	\centering
	\includegraphics[width=1\textwidth]{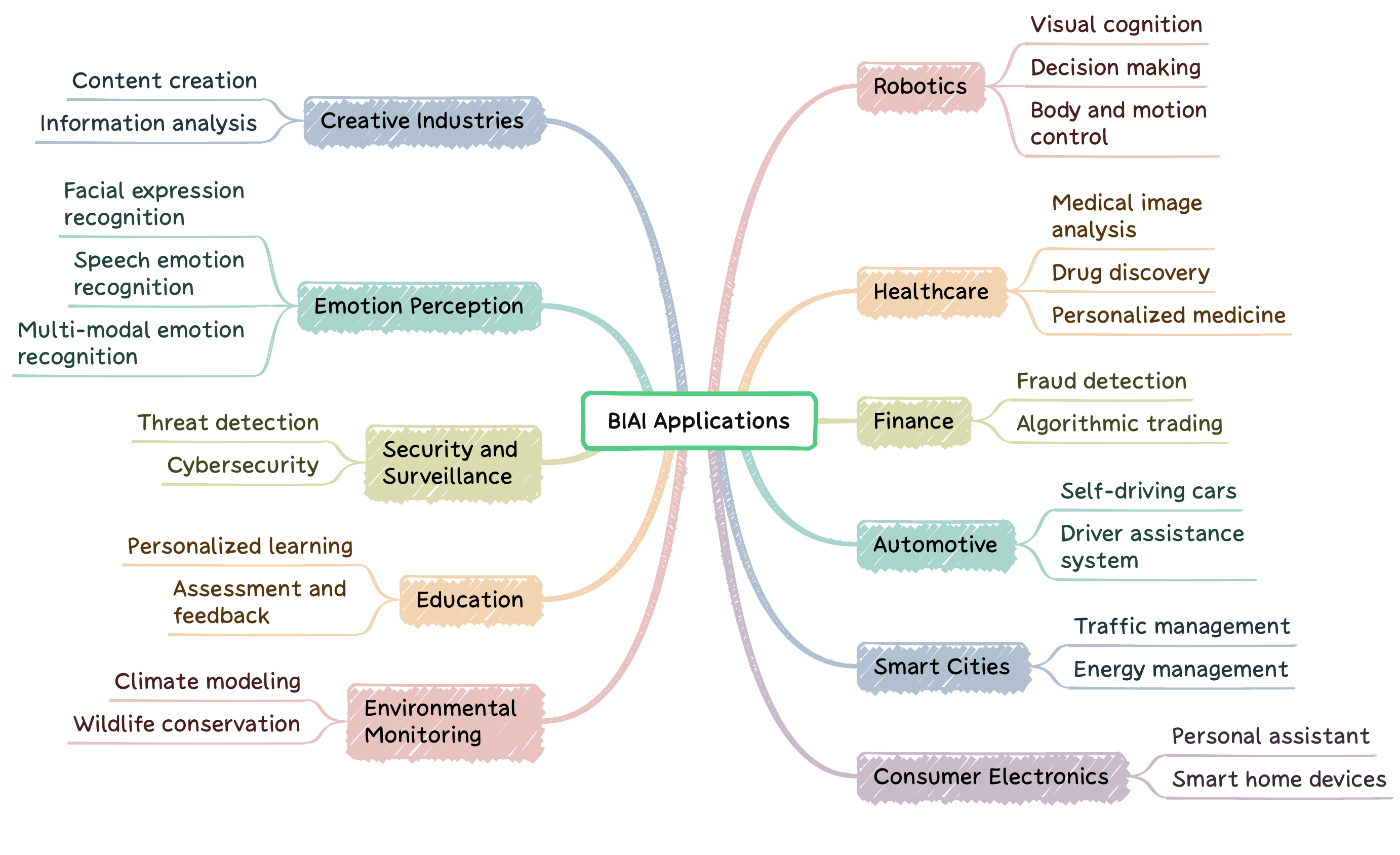}\\
	\caption{BIAI Applications.}\label{app}
\end{figure*}

The applications of BIAI models are vast and diverse, spanning across industries and disciplines, and they continue to drive innovation and advancement in a wide range of fields. A summarization of these applications is given in Fig.~\ref{app}. Due to space limitations, we only highlight four representative applications in this section, demonstrating how popular tasks within these areas are effectively addressed by incorporating BIAI.

\subsection{Robotics}
\begin{figure*}[htb]
	\centering
	\includegraphics[width=1\textwidth]{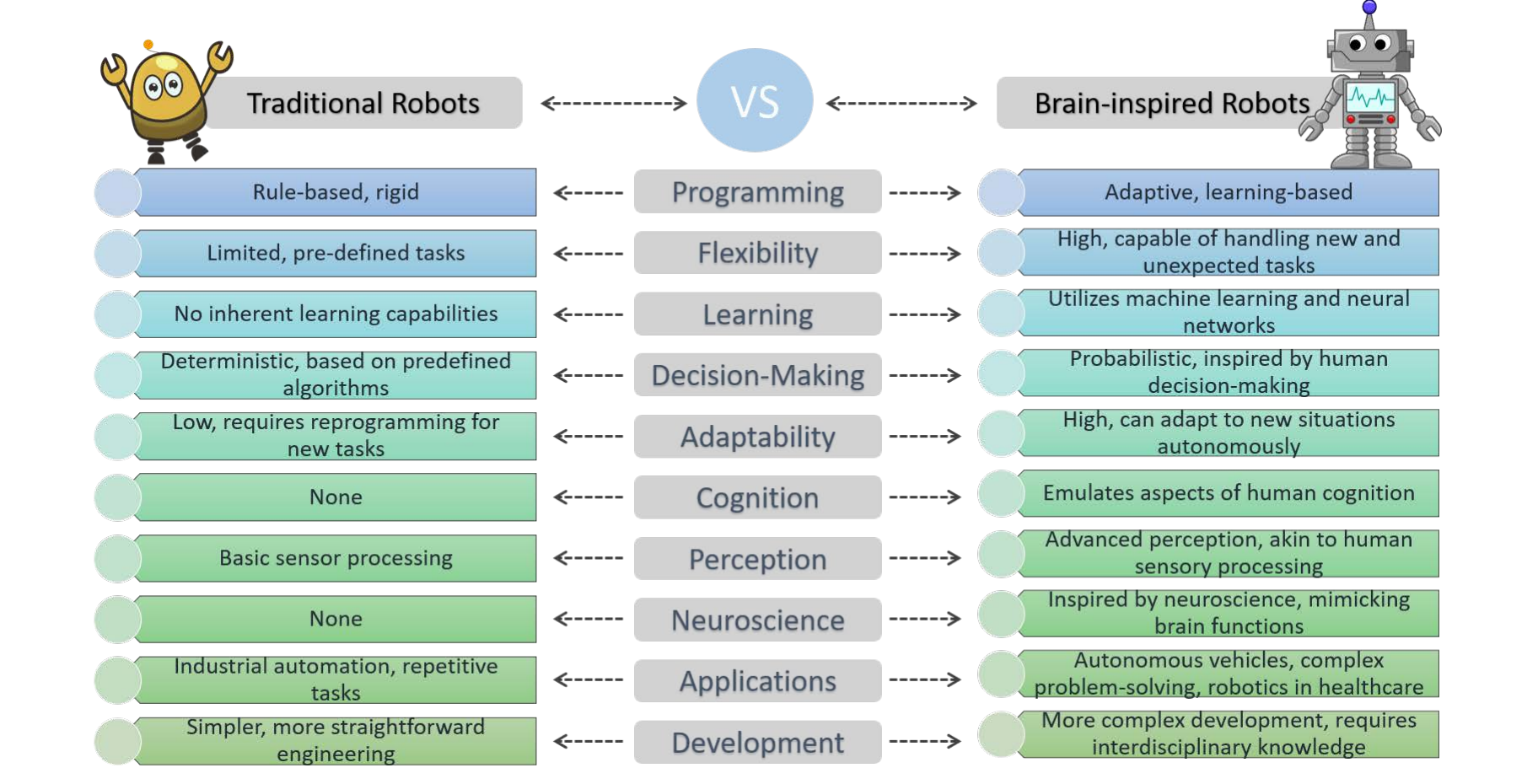}\\
	\caption{Differences between traditional robots and brain-inspired robots.}\label{versus}
\end{figure*}

Traditional AI models often struggle to make accurate and rapid decisions in complex environments, especially for tasks involving robotic movements and dexterous manipulations. Additionally, current robotics research often focuses on specific tasks, limiting broader applications, such as precise manipulation or deep collaboration between humans and robots. These challenges hinder the broader application of robotics in various fields. To overcome these challenges, researchers are turning to the brain for inspiration, leading to the development of brain-inspired intelligent robots. By mimicking the brain's neural networks, robots can perceive, learn, and interact with their environment more naturally, paving the way for more human-like robotic capabilities~\cite{qiao2021survey}. A concret comparison between traditional robots and brain-inspired robots is shown in Fig.~\ref{versus}.

\subsubsection{Visual cognition}
Traditional deep-learning-based models for visual cognition face several challenges. Firstly, they require extensive datasets for training, which are often unavailable in real-world applications~\cite{ren2021deep}. Additionally, these models are typically sensitive to disturbances. Even minor noise in input image data can lead to significant deviations in the output~\cite{hou2021a3graph}. Furthermore, these models operate as black boxes, with their internal workings largely unexplained. This makes it very hard, if not impossible, to understand the decision-making processes of models. Lastly, these visual models are usually designed for certain specific tasks, and hence they may lack the flexibility to quickly adapt to new tasks.

\subsubsection{Decision making}
In decision-making, key challenges include autonomously acquiring environmental knowledge and making swift, accurate decisions in complex scenarios. Advances in this field could significantly enhance the efficiency and precision of robotic movements and complex manipulations, impacting intelligent manufacturing and everyday life. Recently, progress in integrating AI, robotics, and neuroscience has led to the development of various learning-based decision-making techniques that have demonstrated impressive performance in autonomously acquiring robotic knowledge and skills~\cite{hang2023brain}. Despite these advancements, several common issues persist, such as low learning efficiency, difficulty in developing goal-oriented strategies, and limited generalization capability.

\subsubsection{Body and motion control}
In the realm of robotic body and motion control, current joint-link robots are designed to emulate human appearance and functions. However, they fall short in achieving human-like manipulation and interaction capabilities. Conversely, musculoskeletal robots replicate the human skeletal structure, joints, muscles, and the interactions between these components. This design offers greater flexibility, compliance, robustness, safety, and adaptability, making these robots more promising for human-like manipulation and interaction \cite{qiao2021survey}. Nonetheless, the high redundancy, coupling, and nonlinearity inherent in musculoskeletal robots introduce significant challenges in motion control. To address these challenges, various methods grounded in control theory and AI have been developed \cite{taylor2021active}. Despite these advancements, musculoskeletal robots still encounter limitations in movement precision, motion generalization, and rapid responsiveness.

In summary, BIAI models enhance robotic systems in manufacturing environments by improving dexterity, efficiency, and adaptability. They enable robots to perform complex tasks such as visual cognition, decision making, and body and motion control.
\subsection{Healthcare}
BIAI in healthcare utilizes computational models and algorithms to improve medical diagnostics, treatment, and patient care. These sophisticated AI systems can process large volumes of medical data, identify patterns, and make highly accurate predictions, assisting in early disease detection and the development of personalized treatment plans. Here, we introduce how BIAI can be applied into promoting the development of medical image analysis, drug discovery, and personalized medicine.

\subsubsection{Medical image analysis}
BIAI holds the potential to revolutionize multiple facets of the healthcare system. One significant application is in medical image analysis, which includes tasks such as image segmentation, diagnosis, and synthesis \cite{zhou2023deep}. Medical image segmentation is particularly vital for quantitative image analysis, as it delivers essential information like the location and size of lesions. CNNs and their variants are highly effective in extracting meaningful features for semantic segmentation tasks~\cite{suganyadevi2022review}. However, despite their success, medical image segmentation still encounters challenges, including limited labeled data, noise, and annotation uncertainty. These issues could be addressed by incorporating BIAI techniques such as FSL.

In healthcare, medical images such as MRI (magnetic resonance imaging), CT (computed tomography), Ultrasound, and X-Rays can be utilized in AI models to create computer-aided diagnosis (CAD) systems. These systems act as a supplementary tool to assist radiologists in interpreting images for various conditions, including skin cancer, brain tumors, and diabetic retinopathy. CNNs have the capability to analyze thousands or even millions of cases, surpassing the review and memorization capacity of human experts. Research indicates that radiologists' accuracy improves significantly when utilizing CAD systems~\cite{doi2015historical}. However, despite the high expectations for AI's accuracy and efficiency in medicine, integrating new CAD tools into clinical practice poses several challenges, such as ensuring that these tools present recommendations intelligently, explain their reasoning, and correlate findings with patients' medical conditions. Ideally, CAD systems should also provide further explanations if clinicians have questions about the recommendations. Incorporating BIAI models into CAD systems can result in the creation of explainable AI models that provide clarity on their decision-making processes. When these tools are properly developed, validated, and implemented, they can significantly enhance clinicians' abilities, leading to increased accuracy, more efficient workflows, and improved patient care.

Despite the extensive use of medical images for diagnosis, treatment planning, and prognostic analysis, trade-offs between resolution, noise, and acquisition time remain a significant challenge~\cite{brown2014magnetic}. For example, higher resolution MRI scans provide more detail but require longer scan times~\cite{sui2019isotropic}. To address this, image synthesis techniques have emerged as a key area of research in medical image computing and imaging physics~\cite{shi2015lrtv}. GANs and their variants, comprising a generator and discriminator, have shown promise in creating realistic medical images, potentially improving image quality and reducing acquisition times.

\subsubsection{Drug discovery}
Drug discovery encompasses the processes involved in creating and bringing novel drugs to market. AI is accelerating these processes by improving efficiency and speed. Specifically, BIAI models show promise in tackling various challenges in this field~\cite{chen2018rise}. One major issue is that traditional AI models often require extensive datasets for effective training. Inspired by the brain's ability to learn from a few examples, matching networks~\cite{vinyals2016matching} leverage auxiliary data to improve model performance with minimal data, representing a form of one-shot learning. To tackle the issue of limited data, Altae et al.~\cite{altae2017low} successfully applied an LSTM model to a small training set, achieving promising results with chemoinformatics datasets.

Another significant challenge is the lack of explainability in AI systems used for drug discovery. AI systems must be transparent to prevent harmful outcomes~\cite{manresa2021advances}. Understanding the reasoning behind AI decisions is crucial for trust and accountability. To enhance the transparency and reliability of these systems, interfaces have been developed to present justifications for the AI's decisions to users~\cite{kirbouga2023ignition}. Additionally, for molecule generation, generative models based on graph learning have been developed to create molecules with specific desired properties~\cite{xu2019incorporating,ren2023graph}.

\subsubsection{Personalized medicine}
Personalized medicine is transforming healthcare by tailoring treatments to individual patients, offering the potential to revolutionize disease management~\cite{papadakis2019deep}. Traditional AI models typically use rule-based algorithms or statistical methods to support clinical decisions based on broad population data. In contrast, BIAI models can adapt recommendations in real-time based on specific patient data and feedback, offering more personalized and relevant guidance. These BIAI-driven systems integrate individual genomic information, clinical metrics, treatment protocols, and real-time health data to customize treatment recommendations. Traditional AI approaches, often criticized for their lack of interpretability and causality, may create uncertainty or hesitation in their predictions, especially in critical real-world scenarios~\cite{schork2019artificial}. To address these concerns, digital twin technology is emerging as a solution, creating virtual replicas of patients, medical devices, and healthcare systems~\cite{garg2020digital}. By integrating diverse data sources, digital twins can simulate and analyze healthcare processes in real-time, identifying issues and optimizing treatment strategies~\cite{uysal2024explainability}.

Overall, BIAI has the potential to significantly advance healthcare by enhancing diagnostic precision, speeding up drug discovery, and facilitating personalized treatment strategies. As technology progresses, BIAI is anticipated to become increasingly crucial in other healthcare-related fields, including health monitoring and management, as well as clinical research and trials.

\subsection{Emotion Perception}
Emotion perception refers to the process by which individuals recognize and interpret the emotional states of others. It involves the ability to detect and understand emotions from various cues such as facial expressions, vocal tones, and multimodal emotion with contextual information. Emotion perception is not just limited to human interaction but is also a critical component in developing emotionally intelligent machines and systems, such as virtual assistants and robots, that can interact more naturally and effectively with humans. BIAI models are recently used in emotion perception to enhance the ability of machines to recognize and respond to human emotions from the perspectives of vision, voice, and multimodal emtion with contextual information~\cite{xu2024beyond}.
\subsubsection{Facial expression recognition}
Deep learning has propelled facial expression recognition into a leading field of computer vision research~\cite{patel2020facial}. Particularly, BIAI models employ hierarchical processing, similar to the organization of neural circuits in the brain, to analyze facial expressions at multiple levels of abstraction. This allows the models to capture both global facial configurations and fine-grained details, such as changes in muscle movements and microexpressions. Traditional AI models may rely on static feature extraction techniques or predefined rules, limiting their ability to adapt to variations in facial expressions or environmental conditions. BIAI models, on the other hand, excel at continuous learning and adaptation. They refine their understanding of data through iterative training and feedback, demonstrating superior flexibility compared to traditional models. Besides, facial expressions are just one modality used to recognize human behavior. The incorporation of additional data types, such as infrared images, 3D models, and physiological measurements, is an expanding research area because of the valuable complementary information they offer~\cite{mittal2020m3er}.
\subsubsection{Speech emotion recognition}
As for voice emotion recognition, the primary challenge is that background sounds and ambient noise can interfere with the clarity of speech, making it difficult for models to accurately detect emotions~\cite{singh2022systematic}. To solve this problem, BIAI models can dynamically focus on important parts of the speech signal, reducing the impact of overlapping speech and environmental noise, which is inspired by the brain’s selective attention capabilities~\cite{jiao2020real}. Different from facial emotion analysis, emotional expressions can vary significantly across languages, posing a challenge for multilingual or cross-linguistic speech emotion recognition systems~\cite{jahangir2021deep}. BIAI models can be designed to adapt to cultural differences in emotional expression by learning culture-specific patterns and incorporating cross-cultural training data. Another difficulty is relevant with context. Factors such as the topic of conversation, relationship between speakers, and situational context can influence emotional expression. Inspired by the brain’s memory networks, BIAI can employ RNNs and LSTM networks to retain contextual information over time, enhancing the understanding of situational factors~\cite{lian2021ctnet}.

\subsubsection{Multi-modal emotion recognition}
Generating human emotions is inherently complex and subjective, influenced by cognitive processes, biological understanding, and psychophysiological phenomena. As a result, relying solely on traditional machine learning methods for emotion recognition is challenging. Instead, there is a need to create domain-independent, adaptive, and deep machine learning techniques that are transferable for affective computing, leveraging speech, text, and physiological data~\cite{zhang2020emotion}. 
Combining data from various modalities into a cohesive model is complex, because we have to ensure that data from different modalities are temporally aligned and synchronized. BIAI models can utilize neural mechanisms that align and synchronize multimodal inputs, similar to how the brain synchronizes auditory and visual information~\cite{hou2022multi}. 
Different from BIAI models, traditional AI models often handle each modality separately and then integrate the results using predefined fusion methods. In contrast, BIAI models mimic the brain's ability to seamlessly integrate multimodal information, leading to more robust and contextually rich emotion recognition. Moreover,  BIAI models can consider contextual cues, such as the speaker's identity, gender, age, cultural background, interpersonal relationships, and situational context, to better interpret and contextualize emotional expressions~\cite{chen2021learning,mittal2020m3er}. 

In conclusion, BIAI models enhance emotion perception by mimicking the brain's adaptive learning, integration of multiple sensory inputs, and contextual understanding, making it more effective than traditional AI models in recognizing and responding to human emotions.

\subsection{Creative Industries}
Creative content creation includes different perspectives of artistic works, such as music compositions, paintings, and storytelling~\cite{anantrasirichai2022artificial}. BIAI models like GANs and VAEs can learn artistic styles by analyzing vast datasets. These models can then generate new artworks that mimic the styles of famous painters or create entirely original pieces. 
\subsubsection{Content creation}
Music composition, or generation, involves creating sequences of pitches and rhythms in specific patterns. This process demands creativity—a uniquely human ability to understand and produce a diverse range of musical expressions, many of which are entirely novel and previously unexplored~\cite{hernandez2022music}. The tasks involved in music composition include melody generation, multi-instrument arrangement, style transfer, and harmonization. Recent research indicates that transformers and generative models have become prominent neural network architectures in this field~\cite{shehzad2024graph}. The primary challenge in music composition is to produce innovative and engaging works while avoiding cliches. BIAI models, such as those utilizing GANs or RNNs, can generate original music, including melodies, harmonies, and rhythms that might be beyond human imagination, thereby supporting the creative process~\cite{dash2024ai}. Additional challenges in music composition include replicating emotional expression, maintaining structural coherence, and specializing in various genres.

There have been several music composition model generated by AI. For example, Google's Magenta\footnote{https://magenta.tensorflow.org/} focuses on using machine learning to create art and music. It provides tools and models that assist composers in generating new musical ideas and exploring different creative possibilities~\cite{roberts2016interactive}. OpenAI's MuseNet\footnote{https://openai.com/index/musenet/} can generate compositions in various styles and genres by learning from a vast corpus of music. It can create coherent pieces that adhere to the stylistic norms of different genres. NLP models like GPT, are trained on vast corpora of text to understand language semantics, syntax, and context. These models can generate coherent and contextually relevant narratives, poems, and prose passages. AI Dungeon\footnote{https://aidungeon.com/} is an interactive storytelling platform where users can engage in collaborative storytelling with a BIAI-powered dungeon master, generating dynamically evolving narratives in various genres and settings.
\subsubsection{Information analysis}
AI excels at analyzing data to classify content and predict outcomes. This capability is transforming industries. For instance, in advertising, AI helps target audiences effectively. In film, it aids in analysis and content discovery. Journalists leverage AI to find relevant information rapidly~\cite{anantrasirichai2022artificial}.

AI is currently a creative tool, not a competitor. While AI can generate impressive outputs, it's primarily designed to augment human creativity rather than replace it. BIAI models serve as creativity tools, aiding artists, designers, and creators by generating ideas, exploring design variations, and overcoming creative blocks. These tools offer intelligent recommendations, suggestions, and insights to inspire and support the creative process.

Current effective AI often relies on supervised learning, which requires pre-labeled data. This approach is ill-suited for creative tasks lacking clear-cut outcomes~\cite{nixon2020creativity}. Creativity often involves abstract synthesis and experimentation, making data labeling challenging. Unlike humans who learn through diverse experiences, AI struggles to generalize and innovate. To overcome these limitations, future AI models must embrace reinforcement and transfer learning, enabling them to learn from experience and adapt to new challenges~\cite{zhu2023transfer}.

Overall, BIAI is revolutionizing creative content creation by enhancing human creativity with machine intelligence. This fusion enables innovative forms of expression, exploration, and collaboration across art, music, literature, design, and other fields. As BIAI progresses, it has the potential to unlock new dimensions of creativity and expand the limits of artistic expression.

\section{Challenges}~\label{sec5}
Learning from neuroscience to develop BIAI systems presents a multitude of challenges. In this section, we will discuss the most relevant open challenges spanning across several domains, including the complexity of brain structures, limitations in current technology, ethical considerations, and the intrinsic differences between biological and artificial systems.

\textbf{Complexity and Understanding of Brain Function}
Understanding and harnessing brain function for AI development is highly complex due to the brain's intricate structure and processes. With around 86 billion neurons, each connecting to thousands of others, the human brain contains a quadrillion synapses, forming a complex network essential for information processing~\cite{azevedo2009equal}. Modeling such an extensive network with current computational resources presents a substantial challenge. Additionally, the coding schemes used by neurons are not yet fully understood or replicated in BIAI systems, making this an ongoing area of research.

The human brain is an incredibly complex organ orchestrating a wide range of functions essential to human life. These functions can be broadly categorized into cognitive, sensory, motor, emotional, and autonomic processes~\cite{timmann2010human}. Apart from these basic functions that many of them have been applied into existing BIAI models, human brain has higher-order functions including consciousness and creativity. Developing BIAI that can learn and exhibit consciousness like the human brain involves many difficulties, including the defining and understanding of consciousness, the intricate and dynamic nature of brain processes, the need for subjective experience and self-awareness, and the ethical considerations of creating entities that hold consciousness. 

Creativity and imagination of human brain refers to the functions of creative thinking, problem-solving, and imagining future scenarios~\cite{goldberg2018creativity}. Understanding and replicating human creativity poses significant challenges due to its complexity. These challenges span defining and understanding creativity, replicating the complex neural and cognitive mechanisms, generating novel and useful ideas, evaluating and learning from feedback, and addressing emotional, social, ethical, and philosophical issues. Advancing BIAI in this domain requires interdisciplinary research, integrating insights from neuroscience, cognitive science, psychology, and computer science, and addressing both technical and ethical considerations~\cite{ahmad2017cognitive}.

\textbf{Technological and Computational Challenges}
Despite that researchers have mastered abundant neuroscience knowledge, exactly applying these insights into designing specific BIAI models for real-world applications is still a daunting task. When AI learns from the human brain, the technological and computational challenges it faces mainly stem from the complexity of brain structures, the limitations of current technology, and the need for efficient and scalable AI systems.

Emulating the brain’s extensive network of neurons and synapses demands immense computational resources. Current supercomputers can only model a fraction of the human brain's complexity in real-time. Despite this, advancements in benchmarks such as image and text classification have led to increased network complexity, parameter counts, training resource requirements, and prediction latency. For example, Mistral 7B, with its 7 billion parameters, requires millions of dollars for each training iteration~\cite{jiang2023mistral}, not including the costs associated with experimentation and hyperparameter tuning, which are also computationally intensive. Although these models excel at their specific tasks, they may not be efficient enough for direct real-world applications~\cite{menghani2023efficient}. Challenges also arise in training or deploying models for real-time applications on Internet of Things (IoT) and smart devices, where inference occurs directly on the device due to factors like privacy and responsiveness. Thus, optimizing models for these target devices is crucial.

BIAI models often necessitate extensive high-quality data for effective training. For instance, GPT-4, like its predecessors, is trained on datasets containing hundreds of billions to trillions of tokens~\cite{achiam2023gpt}, sourced from a diverse range of texts such as books, websites, and other textual sources. Gemini’s training involves even more extensive multimodal data, including text, images, and audio, amounting to trillions of tokens~\cite{team2023gemini}. Effective multimodal model training depends not only on the quantity of data but also on its variety and richness. This is a stark contrast to human-like learning, where we actively seek information, continually learn, and build on previous experiences. Currently, replicating these capabilities in artificial systems remains a challenge. In the age of big data, algorithms demand ever-larger datasets, while many real-world applications can only work with limited data due to the high costs and time involved in data acquisition~\cite{wang2020generalizing}. Thus, while large BIAI models may take days or weeks to train, the human brain adapts to new environments much more rapidly. This disparity highlights an opportunity to explore how machines can learn more efficiently, akin to human learning.


\textbf{Ethical and Societal Challenges}
Developing BIAI systems involves several ethical and societal challenges. These challenges mainly arise from the potential impact of such technologies on data privacy and fairness~\cite{chen2023ai}.

The continuous advancements of BIAI models across various fields have led to the widespread integration of DNNs into production systems. These advancements are largely driven by the growing availability of large datasets and high computational power. However, these datasets are often crowdsourced and may contain sensitive information, such as medical records, brain imaging data, and behavioral data, raising significant privacy concerns~\cite{vinterbo2004privacy}. 
These data are susceptible to misuse or leak through various vulnerabilities. Even when the cloud provider and communication channels are secure, there remains a risk of inference attacks, where an attacker might deduce properties of the training data or uncover the underlying model architecture and parameters. Therefore, ensuring privacy and security of this data is paramount. 

With the rise of BIAI over the past decades and their widespread adoption in various applications, ensuring safety and fairness has become a critical concern for researchers and engineers in recent years~\cite{dwivedi2023explainable,luo2024algorithmic}.
When BIAI technologies are applied in contexts that impact citizens, it is crucial for companies and researchers to ensure there are no unforeseen social consequences, such as bias against gender, ethnicity, or individuals with disabilities~\cite{caton2024fairness}.
BIAI systems, if trained on biased data, can potentially reinforce and magnify existing biases, which leads to unfair treatment of individuals based on race, gender, socioeconomic status, or other factors. Expecting BIAI systems to provide fair and equitable outcomes for all users is a significant challenge, particularly in areas like healthcare, criminal justice, and hiring and recruitment. Therefore, ensuring that BIAI models are transparent and their decision-making processes are understandable can help identify and mitigate biases.   

\textbf{Interdisciplinary Collaboration}
To conduct cutting-edge scientific research, researchers need opportunities for continuous education and professional development to stay updated on advancements in their primary field as well as interdisciplinary areas. However, effective mentorship that bridges multiple disciplines is often lacking. Therefore, interdisciplinary collaborations are essential for advancing BIAI systems, which involve integrating knowledge and expertise from various fields to address the complex challenges and leverage the full potential of AI. While these collaborations have the potential to drive significant advancements, they also present several challenges~\cite{van2011factors}.

Communication barriers are commonly existing in interdisciplinary collaborations~\cite{brignol2024overcoming}. This is because each discipline has its own specialized language and concepts. Miscommunication can arise when researchers from different fields use the same terms to signify different concepts or different terms to describe the same concept. For instance, terms like "neuron" and "network" in neuroscience might have different connotations or analogs in computer science. Different fields prioritize different research questions and outcomes~\cite{morss2021inter}. For instance, neuroscientists might focus on understanding brain mechanisms, while AI researchers may prioritize developing practical applications. Therefore, the differences in research culture and goals among different disciplines may also lead to failures of collaboration. Moreover, the uneven resource allocation is a major challenge existing in interdisciplinary collaboration. Developing and maintaining shared infrastructure that meets the diverse needs of interdisciplinary teams can be resource-intensive.

Ethical and societal considerations are one of the challenges that should also be critically considered in interdisciplinary collaboration. Different disciplines have varying ethical standards and frameworks, which can complicate the establishment of common ethical guidelines for interdisciplinary research~\cite{hall2017need}. Balancing innovation with ethical considerations requires collaboration among ethicists, social scientists, and technologists, which can be challenging to coordinate. Moreover, due to the complexity and variability of data and theories, integrating findings from different fields into a coherent framework is challenging as well. Moreover, developing frameworks that accommodate the methods and findings of multiple disciplines is essential but difficult. Therefore, effective mentorship that spans multiple disciplines and efficient communications among researchers from different fields are crucial in interdisciplinary collaboration.

\section{Future Directions}~\label{sec6}
In this section, we will summarize the gaps in the BIAI literature to discuss major opportunities for future work. Specifically, we will propose several research directions for BIAI algorithms, including the interdisciplinary collaboration with neuroscience, devising efficient, robust, and responsible AI models, and incorporating consciousness into BIAI models.
\subsection{Integration with Neuroscience}
To design AI models that emulate human thinking and behavior, it is essential to first explore how the human brain processes information and responds to various situations. Gaining a deeper understanding of biological brain functions can significantly contribute to creating intelligent machines. Effectively integrating neuroscience concepts into AI requires a solid grasp of fundamental areas in neuroscience and cognitive science~\cite{ullman2019using}.

The two fields of neuroscience and AI have a long and intertwined history of communication and collaboration. In recent decades, the interaction between these fields has diminished, as both fields have become increasingly complex and their disciplinary boundaries have therefore solidified. However, neuroscience can provide a wide range of inspiration for novel algorithms and architectures. By investigating how humans learn, make decisions, and solve problems, models of cognitive process like memory, attention, perception, and language can be designed. In addition, through studying how information is encoded, stored, and retrieved in the brain, researchers can develop mathematical and computational models of neural circuits and brain functions. Apart from the intuitive knowledge of neuroscience like neural network architechtures, learning, memory and attention mechanism, and sensory processing, there are many other advanced and complex aspects we can learn from neuroscience to solve many opening problems existing in the current AI models~\cite{macpherson2021natural,thompson2021forms}. For instance, examining the impact of emotions on decision-making can lead to the creation of BIAI systems that are capable of understanding and simulating human emotions, resulting in more natural interactions between humans and AI~\cite{martinez2005emotions}. Furthermore, understanding disorders like autism, schizophrenia, and Alzheimer’s can inform the design of BIAI systems that can better understand and potentially assist in diagnosing or treating these conditions~\cite{abd2022performance}. In the future, much work remains to be done to close the current gap between machine intelligence and human intelligence. Besides, our understanding of many brain functions is still incomplete, which limits our ability to accurately model these functions in AI systems. As we strive to bridge this divide, we believe that insights from neuroscience will become increasingly essential.

\subsection{Scalability and Efficiency}
Deep learning research has focused on pushing the boundaries of the field, resulting in notable improvements in benchmarks like image and text classification. However, these advancements often come with increased network complexity, a greater number of parameters, higher training resource requirements, and longer prediction latency, which may limit their efficiency for real-world deployment~\cite{menghani2023efficient}. For instance, some deep learning applications need to function in real-time on IoTs and smart devices, where model inference happens directly on the device. This is crucial for factors like privacy, connectivity, and responsiveness. As a result, designing scalable and efficient models for these specific devices becomes essential.

Scalability and efficiency are two distinct yet interrelated characteristics of AI models~\cite{gill2022ai}. Scalability refers to the ability to handle increased workloads, data volumes, and complexity without significant loss of performance~\cite{mayer2020scalable}. It involves adapting to larger datasets, more complex tasks, and more extensive computational resources. Efficiency, on the other hand, pertains to the model’s capability to execute its tasks using minimal resources, such as computational power, memory, and energy~\cite{menghani2023efficient}. While scalability is about maintaining performance under growing demands, efficiency is about optimizing resource usage to achieve the same or better outcomes.

To improve scalability and efficiency by learning from human brain, we can adopt several strategies. For scalability, the brain's hierarchical and modular architecture can inspire the development of AI models that process information in layers and specialized modules, allowing them to manage increasing complexity and data volume effectively. Techniques such as distributed computing and neuromorphic hardware, which mimic the brain’s parallel processing capabilities, can further enhance scalability. For efficiency, the brain’s energy-efficient processing can be emulated through SNNs~\cite{tavanaei2019deep} and event-driven architectures, which process information only when necessary, reducing computational load. Additionally, incorporating biologically plausible learning rules, like Hebbian learning and synaptic plasticity, can enhance learning processes, boosting both the efficiency and adaptability of BIAI models.

\subsection{Robustness and Resilience}
Enhancing the robustness and resilience of BIAI models is essential for ensuring their reliability and trustworthiness in real-world applications. Robustness in BIAI models usually refers to their ability to perform reliably under a wide range of conditions, including the presence of noise, variations in input data, and potential adversarial attacks~\cite{chander2024toward}. Resilience, on the other hand, pertains to the model's capability to recover from or adapt to failures, disturbances, or changes in the environment~\cite{berger2021survey}. Drawing inspiration from the brain, which exhibits remarkable robustness and resilience through mechanisms like redundancy, plasticity, and fault tolerance, we can significantly enhance these characteristics in BIAI models. For instance, the brain's ability to maintain function despite neuronal damage can inspire the design of BIAI systems with redundant pathways and error-correcting mechanisms, ensuring continuity of operation even when parts of the model fail.

Robust BIAI models are capable of managing noisy, incomplete, or adversarial data while maintaining performance and accuracy despite unforeseen challenges~\cite{wang2021towards}. This is especially critical in high-stake domains such as healthcare, finance, and autonomous systems, where errors can have severe consequences. Resilient BIAI systems can adapt to new and changing conditions, offering greater flexibility and durability over time. By improving robustness and resilience, we can develop BIAI systems that are not only more secure and dependable but also effective in diverse and dynamic environments, leading to wider and more impactful adoption of BIAI technologies.

To improve robustness and resilience in BIAI models, techniques such as regularization, dropout, and adversarial training can be employed. Regularization techniques like L2 norm can prevent overfitting, thereby enhancing the model's ability to generalize to new data. Mimicking the brain's synaptic pruning process, dropout randomly deactivates neurons during training, which helps in creating a more robust network~\cite{zunino2021excitation}. Adversarial training strengthens BIAI models against malicious attacks by exposing them to intentionally distorted inputs during training~\cite{chakraborty2021survey}. Additionally, incorporating mechanisms for continual learning and adaptation, similar to the brain's plasticity, allows the model to adjust to new information and recover from errors, further enhancing its resilience. By embedding these brain-inspired strategies, BIAI models can achieve higher levels of reliability and adaptability in real-world applications.

\subsection{Responsible BIAI}
As we progress towards creating highly effective BIAI models, it is of significant importance to ensure that AI researchers can design more responsible models that align with ethical standards and societal considerations as humans do. Ensuring fairness and transparency is necessary to prevent biases and build trust~\cite{mehrabi2021survey}. In other words, the BIAI models should be capable of explaining their decision-making processes alongside delivering results. Additionally, it is important to inform the public about the limitations of BIAI, despite its advancements, and encourage a balanced and critical approach to its application.

Fairness in BIAI involves ensuring that the models do not produce biased outcomes that could disadvantage certain groups or individuals~\cite{liu2022trustworthy,ijcai2024p50,luo2024algorithmic}. This challenge often stems from biased datasets, where historical prejudices and inequalities are embedded. If a BIAI model is trained on unbalanced dataset that reflects past discriminatory practices, it may perpetuate these biases in its recommendations. It seems that the most intuitive solution to mitigate bias is to use diverse and representative datasets during training, but the process of data clearing in a large-scale dataset is usually time-consuming, if not impossible. Actually, the problem of bias or discrimination is inevitable in the real world, which has been widely and commonly existing in human society for a long time. To cope with this problem, the human brain usually adopts cognitive, neural, and social strategies to mitigate bias. Inspired by this process, understanding these mechanisms can provide insights into how we can avoid biased results in BIAI systems. Besides, regular audits and bias detection mechanisms should be established to identify and address any potential unfair biases in BIAI models~\cite{guo2021detecting}.

Transparency in BIAI refers to the ability of the model to explain its decision-making process~\cite{banafa2024transformative}. This is essential for fostering trust and ensuring accountability. When a BIAI system makes a decision especially in critical areas, it is important for users to understand the reasoning behind its conclusions. The ability of explaining the reason of decisions is innate in humans, but rarely accomplished in most BIAI models. This process usually corresponds to the internal logics of human thinking, and this is what BIAI models are missing.  Studying the logics of human brain can help develop techniques such as explainable AI (XAI)~\cite{saeed2023explainable}, which could provide insights into the decision-making processes of the models. This transparency not only aids in trust-building but also helps in identifying and correcting errors or biases in the models.

\subsection{Conscious AI}
Conscious AI not only processes information and makes decisions but also possesses a form of self-awareness, understanding of its own existence, and the ability to experience and respond to emotions. Emotions significantly impact human cognition and decision-making~\cite{xu2024beyond}. Incorporating emotional and motivational factors into BIAI models is complex and requires understanding the underlying mechanisms. Unlike traditional AI systems that focus on classifying, detecting, recognizing, or conversing with humans, conscious AI would have an awareness of its actions and the capacity to understand and simulate human-like emotions and motivations~\cite{esmaeilzadeh2022conscious}. This concept moves beyond simple computational tasks to an AI agent that can engage in complex interactions, demonstrate empathy, and adapt to emotional and social contexts more naturally.

Designing AI consciousness through learning from the human brain involves several difficulties~\cite{esmaeilzadeh2021conscious}. First, it demands a deep comprehension of the neural underpinnings of human consciousness, including how the brain integrates sensory information, maintains self-awareness, and processes emotions. Neuroscientific research, particularly studies on the default mode network of the brain and the role of the prefrontal cortex in conscious thought~\cite{raccah2021does}, can provide valuable insights. BIAI models can be engineered to replicate these processes by employing neural networks that emulate the structure and function of the human brain. Integrating techniques like hierarchical learning and attention mechanisms can enable these systems to display a degree of self-awareness and emotional comprehension.

Conscious AI has a high value of application due to its potential to revolutionize human-computer interactions. In healthcare, conscious AI could provide more empathetic and personalized care, understanding patient emotions and responding appropriately to their needs~\cite{ranjbari2024implications}. In customer service, it could enhance user experience by recognizing and adapting to customer emotions, leading to more effective and satisfying interactions~\cite{esmaeilzadeh2022conscious}. In education, conscious AI could offer personalized learning experiences, adapting to the emotional and motivational states of students to keep them engaged and motivated~\cite{kim2018towards}. Furthermore, in collaborative work environments, AI with a level of consciousness could function as a more intuitive and effective partner, understanding and responding to human emotions and intentions in real-time.

\section{Conclusion}~\label{sec7}
This paper has provided a comprehensive review of the advancements in AI technology inspired by neuroscience. We began by outlining insights gained from studying the human brain and the current state of BIAI. Our review categorizes BIAI research into two main types: those inspired by the brain's physical structure and those modeled after human behavior. We also explored real-world applications of BIAI in various fields, including robotics, healthcare, emotion recognition, and creative industries. Through a critical analysis of current research trends, we have pinpointed key challenges in BIAI and suggested potential research directions for future exploration. This comprehensive review aims to serve as a valuable resource for BIAI researchers, offering a clear guide to the current landscape of BIAI and its core challenges, ultimately fostering meaningful progress in this field.
\bibliographystyle{ACM-Reference-Format}
\bibliography{BIAI}
	
	
\end{document}